\documentclass[11pt, a4paper]{article}

\usepackage{graphicx}
\usepackage{bm}
\usepackage{mathptmx}
\usepackage{amsmath}
\usepackage{tikz}
\usepackage{fullpage}
\usepackage[round]{natbib} 
\usepackage{xcolor}
\usepackage[multiple]{footmisc}
\usepackage{hyperref}
\usepackage{listings}
\usepackage{subcaption}
\usepackage[cmbtt]{bold-extra}

\newcommand{\lambeqlambda}{{$\lambda\hspace{-0.04cm}\textrm{ambeq}$}}
\newcommand{\lambeq}{\texttt{lambeq}}

\newcounter{example}
\newcommand{\examplecaption}{}

\newenvironment{example}[1]
{
  \renewcommand{\examplecaption}{#1}
  \refstepcounter{example}
  \noindent \textit{Example~\theexample.}~\examplecaption \rmfamily   
}
{
  \par\medskip
}

\def\and{
  \hskip 1.5em
}

\pgfdeclarelayer{edgelayer}
\pgfdeclarelayer{nodelayer}
\pgfsetlayers{edgelayer,nodelayer,main}

\tikzstyle{none}=[inner sep=0pt, thick]
\tikzstyle{plain}=[inner sep=0pt, thick]
\tikzstyle{every picture}=[baseline=(current bounding box).east, scale=0.4, node distance=5mm, text height=5pt, text depth=0pt]

\tikzstyle{black_dot}=[fill=black, draw=black, shape=circle, inner sep=0pt, minimum size=0.2cm, text height=2pt, text depth=0pt]
\tikzstyle{white_dot}=[fill=none, draw=black, shape=circle, inner sep=0pt, minimum size=0.2cm]

\tikzstyle{dashed_line}=[dashed, gray, line width=0.25mm]

\newcommand{\ctikzfig}[1]{%
\begin{center}\rm
   
\InputIfFileExists{#1.tikz}{}{\input{.//tikz_figures//#1.tikz}}

\end{center}}

\title{\textbf{lambeq: An Efficient High-Level Python\\ Library for Quantum NLP}}
\author{
Dimitri Kartsaklis\and 
Ian Fan\and 
Richie Yeung\and
Anna Pearson\\
Robin Lorenz\and
Alexis Toumi\and
Giovanni de Felice\\
Konstantinos Meichanetzidis\and
Stephen Clark\and
Bob Coecke\vspace{0.2cm}\\
Cambridge Quantum Computing\\
\textit{17 Beaumont Street, Oxford, OX1 2NA, UK}\vspace{0.2cm} \\
\texttt{\normalsize \{dimitri.kartsaklis;ian.fan;richie.yeung;anna.pearson;}\vspace{-0.1cm}\\
\texttt{\normalsize robin.lorenz;alexis.toumi;giovanni.defelice;}\vspace{-0.1cm} \\
\texttt{\normalsize kmei;steve.clark;bob.coecke\}@cambridgequantum.com} \\
}

\date{}

\begin{document}

\interfootnotelinepenalty=10000 

\maketitle

\begin{abstract}
\noindent
We present \lambeq, the first high-level Python library for Quantum Natural Language Processing (QNLP). The open-source toolkit offers a detailed hierarchy of modules and classes implementing all stages of a pipeline for converting sentences to string diagrams, tensor networks, and quantum circuits ready to be used on a quantum computer. \lambeq~supports syntactic parsing, rewriting and simplification of string diagrams, ansatz creation and manipulation, as well as a number of compositional models for preparing quantum-friendly representations of sentences, employing various degrees of syntax sensitivity. We present the generic architecture and describe the most important modules in detail, demonstrating the usage with illustrative examples. Further, we test the toolkit in practice by using it to perform a number of experiments on simple NLP tasks, implementing both classical and quantum pipelines.
\end{abstract}

\section{Introduction}
\label{sec:intro}

\textit{Quantum Natural Language Processing} (QNLP) is a very young area of research, aimed at the design and implementation of NLP models that exploit certain quantum phenomena such as superposition, entanglement, and interference to perform language-related tasks on quantum hardware. The advent of the first quantum machines, known as \textit{noisy intermediate-scale quantum} (NISQ) computers, has already allowed researchers to make the first small steps towards exploring practical QNLP, by training models and running simple NLP experiments on quantum hardware \citep{meichanetzidis2020grammaraware,lorenz2021qnlp}. Despite the limited capabilities of the current quantum machines, this early work is important in helping us understand better the process, the technicalities, and the unique nature of this new computational paradigm. At this stage, getting more hands-on experience is crucial in closing the gap that exists between theory and practice, and eventually leading to a point where practical real-world QNLP applications will become a reality. 

With that goal in mind, we introduce \lambeq\footnote{Stylised `\lambeqlambda', pronounced ``lambek''. The name is a tribute to mathematician Joachim Lambek (1922-2014), whose seminal work lay at the intersection of mathematics, logic, and linguistics.}\footnote{\url{https://github.com/CQCL/lambeq}}, an open-source, modular, extensible high-level Python library, which provides the necessary tools for implementing a pipeline for experimental QNLP. At a high level, the library allows the conversion of any sentence to a quantum circuit, based on a given compositional model and certain parameterisation and choices of ans\"atze. This first version of the library includes three compositional models, each using syntactic information to a different degree: a bag-of-words model with no syntactic information present, a word-sequence model which respects the order of words, and a fully syntax-based model following the compositional distributional framework from \cite{CoeckeSadrClark2010}, often dubbed DisCoCat\footnote{DIStributional COmpositional CATegorical.}. It should be noted that \lambeq~is extensible, and in practice it can accommodate any compositional model that can encode sentences as  \textit{string diagrams}, \lambeq's native data structure.\footnote{String diagrams in \lambeq~are close to \textit{tensor networks} (a popular structure in quantum physics), but richer from a representation point of view. See Section \ref{sec:string_diagrams} for more details.}
Further compositional models are currently being developed and will be released in the near future as part of a new version of the toolkit.

In general, \lambeq~loosely follows the pipeline employed in the first small-scale NLP experiments on quantum hardware by \cite{meichanetzidis2020grammaraware} and \cite{lorenz2021qnlp}. Figure \ref{fig:gen_pipeline} shows the processing of a sentence in more detail, the main stages of which are the following:

\begin{figure}
\centering
\includegraphics[scale=0.25]{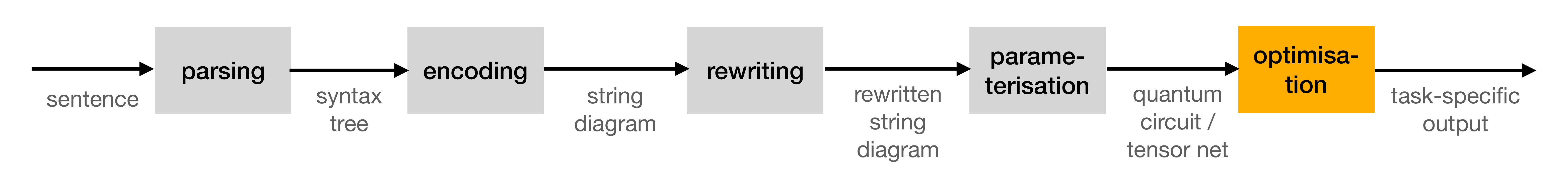}
\caption{Pipeline implemented by \lambeq.}
\label{fig:gen_pipeline}
\end{figure}

\begin{enumerate}
 \item Depending on the selected compositional model, a syntax tree for the sentence might be obtained by calling a statistical CCG\footnote{Combinatory Categorial Grammar \citep{syntactic_process}.} parser. \lambeq~ is equipped with a detailed API that greatly simplifies this process, and is shipped with support for a state-of-the-art parser \citep{yoshikawa:2017acl}.  
 
 \item Internally, the parse tree is converted into a string diagram, following the theory and practice described in \cite{yeung-kartsaklis}. The string diagram can be seen as an abstract representation of the sentence reflecting the relationships between the words as defined by the compositional model of choice, independently of any implementation decisions that take place at a lower level. For storing and manipulating string diagrams, \lambeq~uses as a backend DisCoPy \citep{discopy}, a specialised Python library designed for this purpose.
 
 \item The string diagram can be simplified or otherwise transformed by the application of rewriting rules; these can be used for example to remove specific interactions between words that might be considered redundant for the task at hand, or in order to make the computation more amenable to implementation on a quantum processing unit.
 
 \item\label{last_stage} Finally, the resulting string diagram can be converted into a concrete quantum circuit (or a tensor network in the case of a ``classical'' experiment\footnote{In this paper we loosely use the term ``quantum experiment/pipeline'' to refer to an experiment or pipeline that uses quantum circuits on quantum or classical hardware. This is as opposed to a ``classical experiment/pipeline'', which mainly involves training tensor networks on classical hardware.}), based on a specific parameterisation scheme and concrete choices of ans\"atze. \lambeq~features an extensible class hierarchy containing a selection of pre-defined ans\"atze, appropriate for both classical and quantum experiments.
 
 \end{enumerate}
 
After Step \ref{last_stage}, the output of the pipeline (quantum circuit or tensor network) is ready to be used for training. In the case of a fully quantum pipeline, the quantum circuit will be processed by a quantum compiler and subsequently uploaded onto a quantum computer, while in the classical case the tensor network will be passed to an ML or optimisation library, such as PyTorch\footnote{\url{https://pytorch.org}} or JAX\footnote{\url{https://github.com/google/jax}}. This first version of \lambeq~does not include any optimisation or training features of its own, which is one of our future goals (for more details see Section \ref{sec:future_work}).

We demonstrate the usage of the toolkit in practice by performing classical and quantum experiments on the meaning classification dataset introduced by \cite{lorenz2021qnlp}. Specifically, we implement:

\begin{itemize}
\item a classical experiment, in which the model encodes sentences as tensor networks and is trained using PyTorch;
\item a classical simulation of a quantum pipeline, where sentences are encoded as quantum circuits with JAX used as part of the training backend;
\item a noiseless shot-based quantum simulation experiment executed on \texttt{qiskit}'s\footnote{\url{https://qiskit.org}} Aer simulator using \texttt{tket}\footnote{\url{https://github.com/CQCL/pytket}}, CQC's quantum compiler.
\item a noisy shot-based quantum simulation experiment, again  on \texttt{qiskit}'s Aer simulator using \texttt{tket}.
 
\end{itemize}

The rest of the paper is structured as follows: Section \ref{sec:background} puts this work in context by providing a summary of related research; Section \ref{sec:design_goals} explains \lambeq's design goals; Section \ref{sec:string_diagrams} discusses string diagrams, \lambeq's method for representing sentences; Section \ref{sec:main_components} describes the two external components on which \lambeq~is based, namely the statistical parser and DisCoPy; Section \ref{sec:architecture} outlines the software architecture and discusses the main modules and classes; Section \ref{sec:practice} describes the experiments, and discusses the results; and finally, Section \ref{sec:future_work} summarises our future goals with regard to subsequent versions of the toolkit. 

\section{Background}
\label{sec:background}

Bringing the promise of a substantial leap forward in computational power, quantum computing has generated rapidly growing interest in recent years, with scientists and engineers forming an active and dedicated research community. Promising results and applications in quantum computing can be found in a wide range of topics, such as cryptography \citep{Pirandola_2020}, chemistry \citep{Cao2019}, and biomedicine \citep{8585034}. One of the areas that has attracted a lot of attention is quantum machine learning \citep{schuld2018supervised}, with some work on quantum neural networks; see for example \citep{GUPTA2001355, qnn}. 

On language-related tasks, \cite{bausch2020quantum} employ Grover search to achieve superpolynomial speedups for parsing, while \cite{wiebe2019quantum} present a Fock-space representation of language, which, together with a Harmony optimisation method, allows them to solve NLP problems on a quantum computer. In \cite{gallego2019language}, parse trees are interpreted as information coarse-graining tensor networks where it is also proposed that they can be instantiated as quantum circuits. \cite{RAMESH2003103} provide quantum speedups
for string-matching. There is also work on quantum-inspired classical models incorporating features of quantum theory, such as its inherent probabilistic nature \citep{basile-tamburini-2017-towards} or the existence of many-body entangled states \citep{chen2020quantum}. 

From an experimental NLP point of view, \cite{meichanetzidis2020grammaraware} provided the first proof of concept that practical QNLP is in principle possible in the NISQ era, by training a classifier on a small dataset of 16 sentences. In follow-up work, \cite{lorenz2021qnlp} scaled up the training on datasets of 100-130 sentences, demonstrating convergence of the models and statistically significant results over a random baseline, thus conducting the first complete small-scale NLP experiments on quantum hardware.

The idea of representing language with string diagrams and monoidal categories first appeared in \cite{CoeckeSadrClark2010}, in the context of DisCoCat -- a compositional model of natural language whose mathematical foundations provided a connection to quantum mechanics. In later work, this connection was exploited so that aspects of the DisCoCat model were leveraged to obtain a quadratic speedup \citep{Zeng2016}. Further theoretical work \citep{nearterm_km} lays the foundations for implementations on NISQ devices. 

Finally, while \lambeq~is the first programming toolkit specifically aimed at QNLP research, there are already a few products available for generic quantum machine learning development. Examples include TensorFlow Quantum \citep{tensorflow-quantum}, Google's library for developing hybrid quantum-classical ML models, and Pennylane \citep{pennylane}, a cross-platform Python library for quantum machine learning and automatic differentiation developed by Xanadu.

\section{Design goals}
\label{sec:design_goals}

In this section we briefly discuss the main design goals of the toolkit. \lambeq~is written in Python, which due to its fast prototyping abilities, extensibility, and the support it receives from a large and dedicated community, is currently perhaps the most popular programming language in academia and scientific computing. This choice allows \lambeq~to greatly benefit from Python's open-source ecosystem and the various freely available scientific and numerical libraries such as \texttt{numpy}, \texttt{scipy}, PyTorch, JAX and many more. 

The toolkit is open-source, published under the 
Apache 2.0 licence\footnote{\url{https://www.apache.org/licenses/LICENSE-2.0}}. This allows code transparency, better communication with users and easier reporting of issues. More importantly, members of the expanding QNLP community will be able to make their own contributions and write their own extensions. By open-sourcing \lambeq, we are aiming at active participation of QNLP researchers as part of a focused effort for extending and improving the available functionality.


One of the main design choices for the toolkit was for it to be highly \textit{modular}. We keep the coupling between the various modules as low as possible, so that each of them could be used independently of the others, providing flexibility and extensibility. \textit{Extensibility}, in particular, is another major design goal, and is supported by detailed object-oriented design and a flexible class hierarchy, which allows easy addition of basic components such as compositional models, ans\"atze, and rewrite rules.

Finally, the toolkit was designed with \textit{interoperability} in mind. For example, adding a new parser boils down to encapsulating the appropriate calls into a single wrapper class, for which a detailed API is provided. Further, exporting trees and diagrams into JSON format is inherently supported by both \lambeq~and DisCoPy, simplifying communication with other applications. With regard to quantum hardware, DisCoPy features conversion to \texttt{tket} format, a compiler for optimising and manipulating platform-agnostic quantum circuits. For experiments on classical hardware, users have the ability to export the diagrams in Google's tensor network format\footnote{\url{https://github.com/google/TensorNetwork}}, and use one of the PyTorch or TensorFlow backends for manipulation and training. 

\section{String diagrams}
\label{sec:string_diagrams}

\subsection{Motivation and connection to tensor networks}
\label{sec:tn_motivation}

``Programming'' a quantum computer requires from developers the ability to manipulate \textit{quantum gates} (which can be seen as the ``atomic'' units of computation in this paradigm) in order to create quantum circuits, which can be further grouped into higher-order constructions. Working at such a low level compares to writing assembly in a classical computer, and is extremely hard for humans -- especially on NLP tasks which contain many levels of abstractions. 

\begin{figure}[!h]
\centering
\begin{subfigure}{0.45\textwidth}
\centering

\begin{tikzpicture}
	\begin{pgfonlayer}{nodelayer}
		\node [style=none] (0) at (-8.5, 2) {};
		\node [style=none] (1) at (-8.5, 3) {};
		\node [style=none] (2) at (-5.5, 3) {};
		\node [style=none] (3) at (-5.5, 2) {};
		\node [style=none] (4) at (-5, 2) {};
		\node [style=none] (5) at (-5, 3) {};
		\node [style=none] (6) at (-1, 3) {};
		\node [style=none] (7) at (-1, 2) {};
		\node [style=none] (8) at (-7, 3.5) {};
		\node [style=none] (9) at (-3, 3.5) {};
		\node [style=none] (10) at (3, 2) {};
		\node [style=none] (11) at (3, 3) {};
		\node [style=none] (12) at (6, 3) {};
		\node [style=none] (13) at (6, 2) {};
		\node [style=none] (14) at (4.5, 3.5) {};
		\node [style=none] (15) at (6.5, 2) {};
		\node [style=none] (16) at (6.5, 3) {};
		\node [style=none] (17) at (9.5, 3) {};
		\node [style=none] (18) at (9.5, 2) {};
		\node [style=none] (19) at (8, 3.5) {};
		\node [style=none] (20) at (-7, 2) {};
		\node [style=none] (21) at (-7, 1.5) {};
		\node [style=none] (22) at (-4.5, 2) {};
		\node [style=none] (23) at (-4.5, 1.5) {};
		\node [style=none] (24) at (-3.5, 2) {};
		\node [style=none] (25) at (-3.5, 1.5) {};
		\node [style=none] (26) at (-2.5, 2) {};
		\node [style=none] (27) at (-2.5, 1.5) {};
		\node [style=none] (28) at (3.75, 2) {};
		\node [style=none] (29) at (3.75, 1.5) {};
		\node [style=none] (30) at (8, 2) {};
		\node [style=none] (31) at (8, 1.5) {};
		\node [style=none] (32) at (5.25, 2) {};
		\node [style=none] (33) at (5.25, 1.5) {};
		\node [style=none] (35) at (-7, 1.5) {};
		\node [style=none] (42) at (-4.5, 1.5) {};
		\node [style=none] (43) at (-3.5, -0.75) {};
		\node [style=none] (44) at (-2.5, 1.5) {};
		\node [style=none] (45) at (-3.5, 1.5) {};
		\node [style=none] (46) at (3.75, 1.5) {};
		\node [style=none] (47) at (5.25, 1.5) {};
		\node [style=none] (48) at (8, 1.5) {};
		\node [style=none] (49) at (-7, 2.5) {\footnotesize{John}};
		\node [style=none] (50) at (-3, 2.5) {\footnotesize{gave}};
		\node [style=none] (51) at (4.5, 2.5) {\footnotesize{a}};
		\node [style=none] (52) at (8, 2.5) {\footnotesize{flower}};
		\node [style=none] (53) at (-0.5, 2) {};
		\node [style=none] (54) at (-0.5, 3) {};
		\node [style=none] (55) at (2.5, 3) {};
		\node [style=none] (56) at (2.5, 2) {};
		\node [style=none] (57) at (1, 3.5) {};
		\node [style=none] (58) at (1, 2) {};
		\node [style=none] (59) at (1, 1.5) {};
		\node [style=none] (61) at (1, 1.5) {};
		\node [style=none] (62) at (1, 2.5) {\footnotesize{Mary}};
		\node [style=none] (63) at (-1.5, 2) {};
		\node [style=none] (64) at (-1.5, 1.5) {};
		\node [style=none] (66) at (-1.5, 1.5) {};
	\end{pgfonlayer}
	\begin{pgfonlayer}{edgelayer}
		\draw (0.center) to (3.center);
		\draw (0.center) to (1.center);
		\draw (2.center) to (3.center);
		\draw (4.center) to (7.center);
		\draw (4.center) to (5.center);
		\draw (6.center) to (7.center);
		\draw (1.center) to (8.center);
		\draw (8.center) to (2.center);
		\draw (5.center) to (9.center);
		\draw (9.center) to (6.center);
		\draw (10.center) to (13.center);
		\draw (10.center) to (11.center);
		\draw (12.center) to (13.center);
		\draw (11.center) to (14.center);
		\draw (14.center) to (12.center);
		\draw (15.center) to (18.center);
		\draw (15.center) to (16.center);
		\draw (17.center) to (18.center);
		\draw (16.center) to (19.center);
		\draw (19.center) to (17.center);
		\draw (20.center) to (21.center);
		\draw (22.center) to (23.center);
		\draw (24.center) to (25.center);
		\draw (26.center) to (27.center);
		\draw (28.center) to (29.center);
		\draw (32.center) to (33.center);
		\draw (30.center) to (31.center);
		\draw [in=-90, out=-90, looseness=1.25] (35.center) to (42.center);
		\draw (45.center) to (43.center);
		\draw [bend right=90, looseness=1.25] (47.center) to (48.center);
		\draw (53.center) to (56.center);
		\draw (53.center) to (54.center);
		\draw (55.center) to (56.center);
		\draw (54.center) to (57.center);
		\draw (57.center) to (55.center);
		\draw (58.center) to (59.center);
		\draw (63.center) to (64.center);
		\draw [bend right=90, looseness=1.25] (66.center) to (61.center);
		\draw [bend right=90] (44.center) to (46.center);
	\end{pgfonlayer}
\end{tikzpicture}}

\caption{}
\label{fig:string}
\end{subfigure}
\begin{subfigure}{0.45\textwidth}
\centering

\begin{tikzpicture}
	\begin{pgfonlayer}{nodelayer}
		\node [style=none] (0) at (-4.25, 0.75) {};
		\node [style=none] (1) at (-4.25, 2.25) {};
		\node [style=none] (2) at (-2, 2.25) {};
		\node [style=none] (3) at (-2, 0.75) {};
		\node [style=none] (4) at (-4.25, -1.25) {};
		\node [style=none] (5) at (-4.25, 0.25) {};
		\node [style=none] (6) at (3, 0.25) {};
		\node [style=none] (7) at (3, -1.25) {};
		\node [style=none] (20) at (-3, 0.75) {};
		\node [style=none] (21) at (-3, 0.25) {};
		\node [style=none] (24) at (-0.75, -1.25) {};
		\node [style=none] (25) at (-0.75, -2) {};
		\node [style=none] (44) at (-3, 1.5) {\footnotesize{John}};
		\node [style=none] (45) at (-0.75, -0.5) {\footnotesize{gave}};
		\node [style=none] (46) at (-1.75, 0.75) {};
		\node [style=none] (47) at (-1.75, 2.25) {};
		\node [style=none] (48) at (0.75, 2.25) {};
		\node [style=none] (49) at (0.75, 0.75) {};
		\node [style=none] (50) at (-0.5, 0.75) {};
		\node [style=none] (51) at (-0.5, 0.25) {};
		\node [style=none] (52) at (-0.5, 1.5) {\footnotesize{Mary}};
		\node [style=none] (53) at (1, 0.75) {};
		\node [style=none] (54) at (1, 2.25) {};
		\node [style=none] (55) at (3, 2.25) {};
		\node [style=none] (56) at (3, 0.75) {};
		\node [style=none] (57) at (2, 0.75) {};
		\node [style=none] (58) at (2, 0.25) {};
		\node [style=none] (59) at (2, 1.5) {\footnotesize{a}};
		\node [style=none] (60) at (3.75, 0.75) {};
		\node [style=none] (61) at (3.75, 2.25) {};
		\node [style=none] (62) at (6.25, 2.25) {};
		\node [style=none] (63) at (6.25, 0.75) {};
		\node [style=none] (64) at (3.75, 1.5) {};
		\node [style=none] (65) at (3, 1.5) {};
		\node [style=none] (66) at (5, 1.5) {\footnotesize{flower}};
	\end{pgfonlayer}
	\begin{pgfonlayer}{edgelayer}
		\draw (0.center) to (3.center);
		\draw (0.center) to (1.center);
		\draw (2.center) to (3.center);
		\draw (4.center) to (7.center);
		\draw (4.center) to (5.center);
		\draw (6.center) to (7.center);
		\draw (20.center) to (21.center);
		\draw (24.center) to (25.center);
		\draw (1.center) to (2.center);
		\draw (5.center) to (6.center);
		\draw (46.center) to (49.center);
		\draw (46.center) to (47.center);
		\draw (48.center) to (49.center);
		\draw (50.center) to (51.center);
		\draw (47.center) to (48.center);
		\draw (53.center) to (56.center);
		\draw (53.center) to (54.center);
		\draw (55.center) to (56.center);
		\draw (57.center) to (58.center);
		\draw (54.center) to (55.center);
		\draw (60.center) to (63.center);
		\draw (60.center) to (61.center);
		\draw (62.center) to (63.center);
		\draw (64.center) to (65.center);
		\draw (61.center) to (62.center);
	\end{pgfonlayer}
\end{tikzpicture}}

\caption{}
\label{fig:tensornet}
\end{subfigure}
\caption{String diagram and corresponding tensor network.}
\label{fig:string_diagram}
\end{figure}
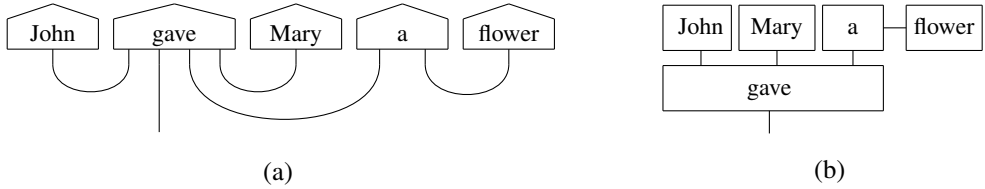

In order to simplify NLP design on quantum hardware, \lambeq~represents sentences as string diagrams (Figure \ref{fig:string}). This choice stems from the fact that a string diagram expresses computations in a \textit{monoidal category}\footnote{For an introduction to monoidal categories, see \cite{baez2010physics,CatsII}.}, an abstraction well-suited to model the way a quantum computer works and processes data. In fact, monoidal categories have previously been used to recast the entire framework of quantum mechanics at a higher conceptual level, forming the field that is now known as \textit{categorical quantum mechanics} \citep{abramsky2004}. From a more practical point of view, a string diagram can be seen as an enriched \textit{tensor network} \citep{tensor_nets,pestun2017tensor} (Figure \ref{fig:tensornet}), a mathematical structure with many applications in quantum physics. Compared to tensor networks, string diagrams have some additional convenient properties, for example they respect the order of words, and allow easy rewriting/modification of their structure (see for example Section \ref{sec:rewriting}).

String diagrams and tensor networks constitute an ideal abstract representation of the compositional relations between the words in a sentence, in the sense that they remain close to quantum circuits, yet are independent of any low-level decisions (such as choice of quantum gates and construction of circuits representing words and sentences) that might vary depending on design choices and the type of quantum hardware that the experiment is running on. More information regarding the diagrammatic representation of quantum processes can be found in \cite{Coecke_Kissinger_2017_TextBook}.

\subsection{Pregroup grammars}
\label{sec:pregroup_grammars}

\lambeq's string diagrams are equipped with types, which show the interactions between the words in a sentence according to the \textit{pregroup grammar} formalism \citep{lambek}. In a pregroup grammar, each type $p$ has a left ($p^l$) and a right ($p^r$) adjoint, for which the following hold:

\begin{equation}
p^l \cdot p \to 1 \to p \cdot p^l \quad\quad\quad\quad p \cdot p^r \to 1 \to p^r \cdot p
\end{equation}

When annotated with pregroup types, the diagram in Figure \ref{fig:string} takes the following form:

\ctikzfig{john-types}

Note that each wire in the sentence is labelled with an atomic type or an adjoint. In the above, $n$ corresponds to a noun or a noun phrase, and $s$ to a sentence. The adjoints $n^r$ and $n^l$ indicate that a noun is expected on the left or the right of the specific word, respectively. Thus, the composite type $n \cdot n^l$ of the determiner ``a'' means that it is a word that expects a noun on its right in order to return a noun phrase. 

The transition from pregroups to vector space semantics is achieved by a mapping that sends atomic types to vector spaces ($n$ to $N$ and $s$ to $S$) and composite types to tensor product spaces (e.g. $n^r\cdot s \cdot n^l \cdot n^l$ to $N\otimes S\otimes N\otimes N$) \citep{reasoning_2016}. Therefore, each word can be seen as a specific state in the corresponding space defined by its grammatical type, i.e. a tensor, the order of which is determined by the number of wires emanating from the corresponding box. The cups ($\cup$) denote tensor contractions. A concrete instantiation of the diagram requires the assignment of dimensions (which in the quantum case amounts to fixing the number of qubits) for each vector space corresponding to an atomic type. More details about parameterisation can be found in Section \ref{sec:parameterisation}. 

\section{Main components}
\label{sec:main_components}

\begin{figure}[!h]
\centering
\includegraphics[scale=0.3]{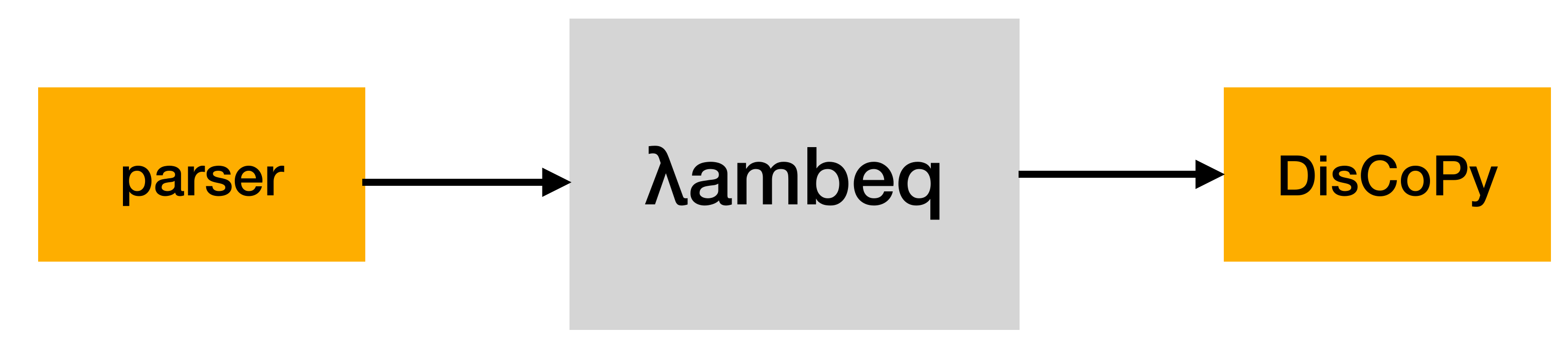}
\caption{\lambeq~as middleware.}
\label{fig:middleware}
\end{figure}

At the highest possible level, \lambeq~can be loosely seen as a case of \textit{middleware}, sitting in between a statistical parser and DisCoPy, as shown in Figure \ref{fig:middleware}. In practice, though, its purpose greatly exceeds the strict definition of middleware as a mere means of communication between applications, since it includes extensive functionality of its own, designed to support language-related tasks. Furthermore, DisCoPy supports the entire range of \lambeq's components with low-level functionality and does not merely provide an output format, as might be implied in the figure. The following sections describe in detail the role of the two main components.

\subsection{CCG parser}
\label{sec:parser}

\lambeq's string diagrams are based on a pregroup grammar (Section \ref{sec:pregroup_grammars}) to keep track of the types and the interactions between the words in a sentence. When a detailed syntactic derivation is required (as in the case of DisCoCat), a syntax tree needs to be provided by a statistical parser. However, since the pregroup grammar formalism is not particularly well-known in the NLP community, there is currently no wide-coverage pregroup parser that can automatically provide the syntactic derivations. To address this problem, \cite{yeung-kartsaklis} recently provided a functorial passage from a derivation in the closest alternative grammar formalism, namely Combinatory Categorial Grammar (CCG) \citep{syntactic_process}, to a string diagram which faithfully encodes the syntactic structure of the sentence in a pregroup-like form. Due to the availability of many robust CCG parsing tools (for example, see \cite{clark-curran-2007-wide}) this allowed the conversion of large corpora with sentences of arbitrary length and syntactic structure into pregroup and DisCoCat form, solving a long-standing problem.\footnote{As an example, the DisCoCat version of the book ``Alice in Wonderland'' can be found at \url{https://qnlp.cambridgequantum.com/downloads.html}; see also Figure \ref{fig:alice}.}

\lambeq~does not use its own statistical CCG parser, but instead it applies the approach of \citeauthor{yeung-kartsaklis} to implement a detailed interface that allows connection to one of the many external CCG parsing tools that are currently available. By default, \lambeq~is shipped with support for \textit{DepCCG} \citep{yoshikawa:2017acl}, a state-of-the-art efficient parser which comes with a convenient Python interface. 

\subsection{DisCoPy}
\label{sec:discopy}

While the parser provides \lambeq's input, DisCoPy\footnote{\url{https://github.com/oxford-quantum-group/discopy}} \citep{discopy} is \lambeq's underlying engine, the component where all the low-level processing takes place. At its core, DisCoPy is a Python library that allows computation with monoidal categories. The main data structure is that of a \textit{monoidal diagram}, or string diagram, which is the format that \lambeq~uses internally to encode a sentence. DisCoPy makes this easy, by offering many language-related features, such as support for pregroup grammars and functors for implementing compositional models such as DisCoCat. Furthermore, from a quantum computing perspective, DisCoPy provides abstractions for creating all standard quantum gates and building quantum circuits, which are used by \lambeq~in the final stages of the pipeline in Figure \ref{fig:gen_pipeline}.

\section{Detailed architecture}
\label{sec:architecture}

\begin{figure}
\centering
\includegraphics[scale=0.27]{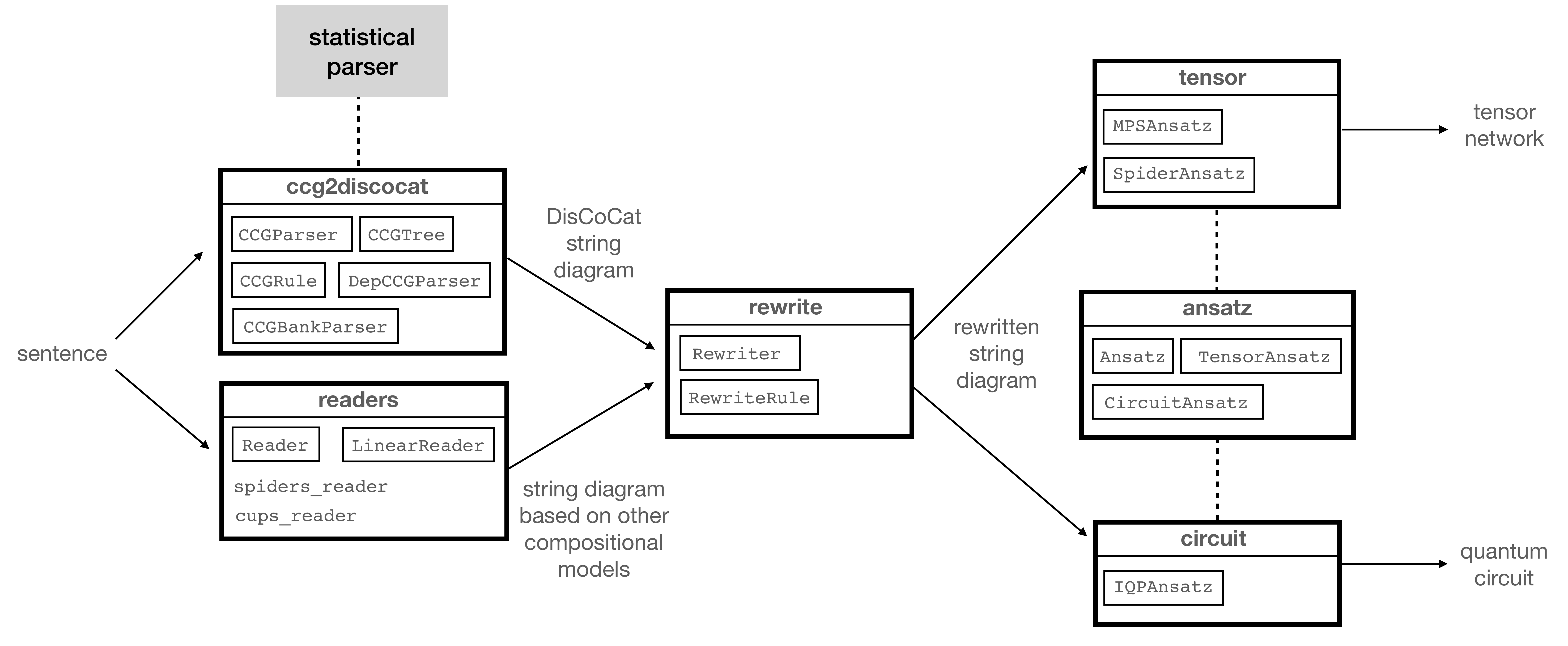}
\caption{\lambeq's architecture and main modules.}
\label{fig:architecture}
\end{figure}

Both the statistical parser and DisCoPy are integrated seamlessly into a unified architecture (Figure \ref{fig:architecture}), implementing all stages of the generic QNLP pipeline, as was given in Figure \ref{fig:gen_pipeline}. In the following sections we describe each in detail, introducing all important \lambeq~modules along the way. More details about \lambeq's API can be found in the toolkit's documentation,\footnote{\lambeq~documentation is available at \url{cqcl.github.io/lambeq}. DisCoPy documentation and usage examples can be found at \url{https://discopy.readthedocs.io/}.} along with extensive usage examples in the form of Jupyter notebooks.

\subsection{Sentence input}
\label{sec:sentence_input}

The first part of the process in \lambeq, given a sentence, is to convert it into a string diagram. In order to obtain a DisCoCat-like output, we first use the \texttt{ccg2discocat} module, which, in turn,  calls the parser, obtains a CCG derivation for the sentence, and converts it into a string diagram. Example \ref{ex:ccg2discocat} uses the default DepCCG parser in order to produce the DisCoCat diagram of Figure \ref{fig:string} for the sentence ``John gave Mary a flower''.\footnote{To see the code examples, we refer the reader to the appendix.} 
Other external parsers can be made available to \lambeq~by extending  the \texttt{CCGParser} class in order to create a wrapper subclass that encapsulates the necessary calls and translates the respective parser's output into \texttt{CCGTree} format. 

DisCoCat is not the only compositional model that \lambeq~supports. In fact, any compositional scheme that manifests sentences as string diagrams/tensor networks can be added to the toolkit via the module \texttt{readers}. For example, the \texttt{spiders\_reader} instance of the  \texttt{LinearReader} class represents a sentence as a ``bag-of-words'', composing the words using a \textit{spider} (a commutative operation), while \texttt{cups\_reader} composes words in sequence, from left to right, generating a ``tensor train'' (Example \ref{ex:readers}). 

\begin{figure}[!h]
\centering
\begin{subfigure}{0.45\textwidth}
\centering

\begin{tikzpicture}
	\begin{pgfonlayer}{nodelayer}
		\node [style=none] (0) at (-8.5, 2) {};
		\node [style=none] (1) at (-8.5, 3) {};
		\node [style=none] (2) at (-5.5, 3) {};
		\node [style=none] (3) at (-5.5, 2) {};
		\node [style=none] (8) at (-7, 3.5) {};
		\node [style=none] (20) at (-7, 2) {};
		\node [style=none] (21) at (-7, 1.5) {};
		\node [style=none] (35) at (-7, 1.5) {};
		\node [style=none] (49) at (-7, 2.5) {\footnotesize{John}};
		\node [style=none] (70) at (-5, 2) {};
		\node [style=none] (71) at (-5, 3) {};
		\node [style=none] (72) at (-2, 3) {};
		\node [style=none] (73) at (-2, 2) {};
		\node [style=none] (74) at (-3.5, 3.5) {};
		\node [style=none] (78) at (-3.5, 2.5) {\footnotesize{gave}};
		\node [style=none] (82) at (-1.5, 2) {};
		\node [style=none] (83) at (-1.5, 3) {};
		\node [style=none] (84) at (1.5, 3) {};
		\node [style=none] (85) at (1.5, 2) {};
		\node [style=none] (86) at (0, 3.5) {};
		\node [style=none] (90) at (0, 2.5) {\footnotesize{Mary}};
		\node [style=none] (94) at (2, 2) {};
		\node [style=none] (95) at (2, 3) {};
		\node [style=none] (96) at (5, 3) {};
		\node [style=none] (97) at (5, 2) {};
		\node [style=none] (98) at (3.5, 3.5) {};
		\node [style=none] (102) at (3.5, 2.5) {\footnotesize{a}};
		\node [style=none] (106) at (5.5, 2) {};
		\node [style=none] (107) at (5.5, 3) {};
		\node [style=none] (108) at (8.5, 3) {};
		\node [style=none] (109) at (8.5, 2) {};
		\node [style=none] (110) at (7, 3.5) {};
		\node [style=none] (114) at (7, 2.5) {\footnotesize{flower}};
		\node [style=none] (115) at (-3.5, 2) {};
		\node [style=none] (116) at (-3.5, 1.5) {};
		\node [style=none] (117) at (-3.5, 1.5) {};
		\node [style=none] (118) at (0, 2) {};
		\node [style={black_dot}] (119) at (0, -0.5) {};
		\node [style=none] (121) at (3.5, 2) {};
		\node [style=none] (122) at (3.5, 1.5) {};
		\node [style=none] (123) at (3.5, 1.5) {};
		\node [style=none] (124) at (7, 2) {};
		\node [style=none] (125) at (7, 1.5) {};
		\node [style=none] (126) at (7, 1.5) {};
		\node [style=none] (127) at (0, -1.5) {};
	\end{pgfonlayer}
	\begin{pgfonlayer}{edgelayer}
		\draw (0.center) to (3.center);
		\draw (0.center) to (1.center);
		\draw (2.center) to (3.center);
		\draw (1.center) to (8.center);
		\draw (8.center) to (2.center);
		\draw (20.center) to (21.center);
		\draw (70.center) to (73.center);
		\draw (70.center) to (71.center);
		\draw (72.center) to (73.center);
		\draw (71.center) to (74.center);
		\draw (74.center) to (72.center);
		\draw (82.center) to (85.center);
		\draw (82.center) to (83.center);
		\draw (84.center) to (85.center);
		\draw (83.center) to (86.center);
		\draw (86.center) to (84.center);
		\draw (94.center) to (97.center);
		\draw (94.center) to (95.center);
		\draw (96.center) to (97.center);
		\draw (95.center) to (98.center);
		\draw (98.center) to (96.center);
		\draw (106.center) to (109.center);
		\draw (106.center) to (107.center);
		\draw (108.center) to (109.center);
		\draw (107.center) to (110.center);
		\draw (110.center) to (108.center);
		\draw (115.center) to (116.center);
		\draw (118.center) to (119);
		\draw (121.center) to (122.center);
		\draw (124.center) to (125.center);
		\draw [bend right=90, looseness=0.50] (35.center) to (126.center);
		\draw [bend right=90] (117.center) to (123.center);
		\draw (119) to (127.center);
	\end{pgfonlayer}
\end{tikzpicture}}

\caption{\texttt{spiders\_reader}}
\end{subfigure}
\begin{subfigure}{0.45\textwidth}
\centering

\begin{tikzpicture}
	\begin{pgfonlayer}{nodelayer}
		\node [style=none] (0) at (-8.5, 2) {};
		\node [style=none] (1) at (-8.5, 3) {};
		\node [style=none] (2) at (-5.5, 3) {};
		\node [style=none] (3) at (-5.5, 2) {};
		\node [style=none] (8) at (-7, 3.5) {};
		\node [style=none] (20) at (-7.75, 2) {};
		\node [style=none] (21) at (-7.75, 1.5) {};
		\node [style=none] (35) at (-7.75, 1.5) {};
		\node [style=none] (49) at (-7, 2.5) {\footnotesize{John}};
		\node [style=none] (67) at (-6.25, 2) {};
		\node [style=none] (68) at (-6.25, 1.5) {};
		\node [style=none] (69) at (-6.25, 1.5) {};
		\node [style=none] (70) at (-5, 2) {};
		\node [style=none] (71) at (-5, 3) {};
		\node [style=none] (72) at (-2, 3) {};
		\node [style=none] (73) at (-2, 2) {};
		\node [style=none] (74) at (-3.5, 3.5) {};
		\node [style=none] (75) at (-4.25, 2) {};
		\node [style=none] (76) at (-4.25, 1.5) {};
		\node [style=none] (77) at (-4.25, 1.5) {};
		\node [style=none] (78) at (-3.5, 2.5) {\footnotesize{gave}};
		\node [style=none] (79) at (-2.75, 2) {};
		\node [style=none] (80) at (-2.75, 1.5) {};
		\node [style=none] (81) at (-2.75, 1.5) {};
		\node [style=none] (82) at (-1.5, 2) {};
		\node [style=none] (83) at (-1.5, 3) {};
		\node [style=none] (84) at (1.5, 3) {};
		\node [style=none] (85) at (1.5, 2) {};
		\node [style=none] (86) at (0, 3.5) {};
		\node [style=none] (87) at (-0.75, 2) {};
		\node [style=none] (88) at (-0.75, 1.5) {};
		\node [style=none] (89) at (-0.75, 1.5) {};
		\node [style=none] (90) at (0, 2.5) {\footnotesize{Mary}};
		\node [style=none] (91) at (0.75, 2) {};
		\node [style=none] (92) at (0.75, 1.5) {};
		\node [style=none] (93) at (0.75, 1.5) {};
		\node [style=none] (94) at (2, 2) {};
		\node [style=none] (95) at (2, 3) {};
		\node [style=none] (96) at (5, 3) {};
		\node [style=none] (97) at (5, 2) {};
		\node [style=none] (98) at (3.5, 3.5) {};
		\node [style=none] (99) at (2.75, 2) {};
		\node [style=none] (100) at (2.75, 1.5) {};
		\node [style=none] (101) at (2.75, 1.5) {};
		\node [style=none] (102) at (3.5, 2.5) {\footnotesize{a}};
		\node [style=none] (103) at (4.25, 2) {};
		\node [style=none] (104) at (4.25, 1.5) {};
		\node [style=none] (105) at (4.25, 1.5) {};
		\node [style=none] (106) at (5.5, 2) {};
		\node [style=none] (107) at (5.5, 3) {};
		\node [style=none] (108) at (8.5, 3) {};
		\node [style=none] (109) at (8.5, 2) {};
		\node [style=none] (110) at (7, 3.5) {};
		\node [style=none] (111) at (6.25, 2) {};
		\node [style=none] (112) at (6.25, 1.5) {};
		\node [style=none] (113) at (6.25, 1.5) {};
		\node [style=none] (114) at (7, 2.5) {\footnotesize{flower}};
		\node [style=none] (115) at (7.75, 2) {};
		\node [style=none] (116) at (7.75, -0.75) {};
		\node [style=none] (118) at (-11, 2) {};
		\node [style=none] (119) at (-11, 3) {};
		\node [style=none] (120) at (-9, 3) {};
		\node [style=none] (121) at (-9, 2) {};
		\node [style=none] (122) at (-10, 3.5) {};
		\node [style=none] (126) at (-10, 2.5) {\footnotesize{$\langle S\rangle$}};
		\node [style=none] (127) at (-10, 2) {};
		\node [style=none] (128) at (-10, 1.5) {};
		\node [style=none] (129) at (-10, 1.5) {};
	\end{pgfonlayer}
	\begin{pgfonlayer}{edgelayer}
		\draw (0.center) to (3.center);
		\draw (0.center) to (1.center);
		\draw (2.center) to (3.center);
		\draw (1.center) to (8.center);
		\draw (8.center) to (2.center);
		\draw (20.center) to (21.center);
		\draw (67.center) to (68.center);
		\draw (70.center) to (73.center);
		\draw (70.center) to (71.center);
		\draw (72.center) to (73.center);
		\draw (71.center) to (74.center);
		\draw (74.center) to (72.center);
		\draw (75.center) to (76.center);
		\draw (79.center) to (80.center);
		\draw (82.center) to (85.center);
		\draw (82.center) to (83.center);
		\draw (84.center) to (85.center);
		\draw (83.center) to (86.center);
		\draw (86.center) to (84.center);
		\draw (87.center) to (88.center);
		\draw (91.center) to (92.center);
		\draw (94.center) to (97.center);
		\draw (94.center) to (95.center);
		\draw (96.center) to (97.center);
		\draw (95.center) to (98.center);
		\draw (98.center) to (96.center);
		\draw (99.center) to (100.center);
		\draw (103.center) to (104.center);
		\draw (106.center) to (109.center);
		\draw (106.center) to (107.center);
		\draw (108.center) to (109.center);
		\draw (107.center) to (110.center);
		\draw (110.center) to (108.center);
		\draw (111.center) to (112.center);
		\draw (115.center) to (116.center);
		\draw (118.center) to (121.center);
		\draw (118.center) to (119.center);
		\draw (120.center) to (121.center);
		\draw (119.center) to (122.center);
		\draw (122.center) to (120.center);
		\draw (127.center) to (128.center);
		\draw [bend right=90, looseness=1.25] (129.center) to (35.center);
		\draw [bend right=90, looseness=1.25] (69.center) to (77.center);
		\draw [bend right=90, looseness=1.25] (81.center) to (89.center);
		\draw [bend right=90, looseness=1.25] (93.center) to (101.center);
		\draw [bend right=90, looseness=1.25] (105.center) to (113.center);
	\end{pgfonlayer}
\end{tikzpicture}}

\caption{\texttt{cups\_reader}}
\end{subfigure}
\end{figure}

Writing new compositional models entails the extension of the \texttt{Reader} class and the implementation of the method \texttt{sentence2diagram} in the subclass using DisCoPy calls. 

Finally, \lambeq~features a \texttt{CCGBankParser} class, which allows conversion of the entire CCGBank corpus \citep{ccgbank} into string diagrams. CCGBank consists of 49,000 human-annotated CCG syntax trees, converted from the original Penn Treebank into CCG form. Having a gold standard corpus of string diagrams allows  various supervised learning scenarios involving automatic diagram generation. Example \ref{ex:ccgbank} shows how to convert Section 00 of CCGBank into string diagrams. 

%

\subsection{Rewriting}
\label{sec:rewriting}

Syntactic derivations in pregroup form can become extremely complicated, which may lead to excessive use of hardware resources and prohibitively long training times. The purpose of the \texttt{rewrite} module is to provide a means to the user to address some of these problems, via rewriting rules that simplify the string diagram. As an example, consider the sentence ``John walks in the park'', the string diagram of which is the following:

\ctikzfig{pp01}

Note that the representation of the preposition is a tensor of order 5 in the ``classical'' case, or a state of 5 quantum systems in the quantum case. Applying the \texttt{prepositional\_phrase} rewriting rule to the diagram (Example \ref{ex:rewrite}) takes advantage of the underlying compact-closed monoidal structure, by using a ``cap'' ($\cap$) to bridge the discontinued subject noun wire within the preposition tensor:

\ctikzfig{pp02}

In the simplified diagram, the order of the preposition tensor is reduced by 2, which at least for a classical experiment is a substantial improvement. This example clearly demonstrates the flexibility of string diagrams compared to simple tensor networks, which was one of the main reasons for choosing them as \lambeq's representation format. \lambeq~comes with a number of standard rewrite rules covering auxiliary verbs, connectors, adverbs, determiners and prepositional phrases (Table \ref{tbl:rewrite}). 

\begin{table}[h!]
\small
\centering
\begin{tabular}{|r|l|}
\hline
\textbf{Rule} & \textbf{Description} \\
\hline\hline
\texttt{auxiliary} & Removes auxiliary verbs (such as ``do'') by replacing them with caps. \\
\texttt{connector} & Removes sentence connectors (such as ``that'') by replacing them with caps. \\
\texttt{determiner} & Removes determiners (such as ``the'') by replacing them with caps. \\
\texttt{postadverb}, \texttt{preadverb} & Simplifies adverbs by passing through the noun wire transparently\\
 & using a cap. \\
\texttt{prepositional\_phrase} & Simplifies the preposition in a prepositional phrase by passing through \\
 & the noun wire transparently using a cap.\\
\hline
\end{tabular}
\caption{Rewriting rules.}
\label{tbl:rewrite}
\end{table}

\subsection{Parameterisation}
\label{sec:parameterisation}

\begin{figure}[b!]
\centering
\includegraphics[scale=0.37, trim=8cm 3cm 5cm 3cm, clip]{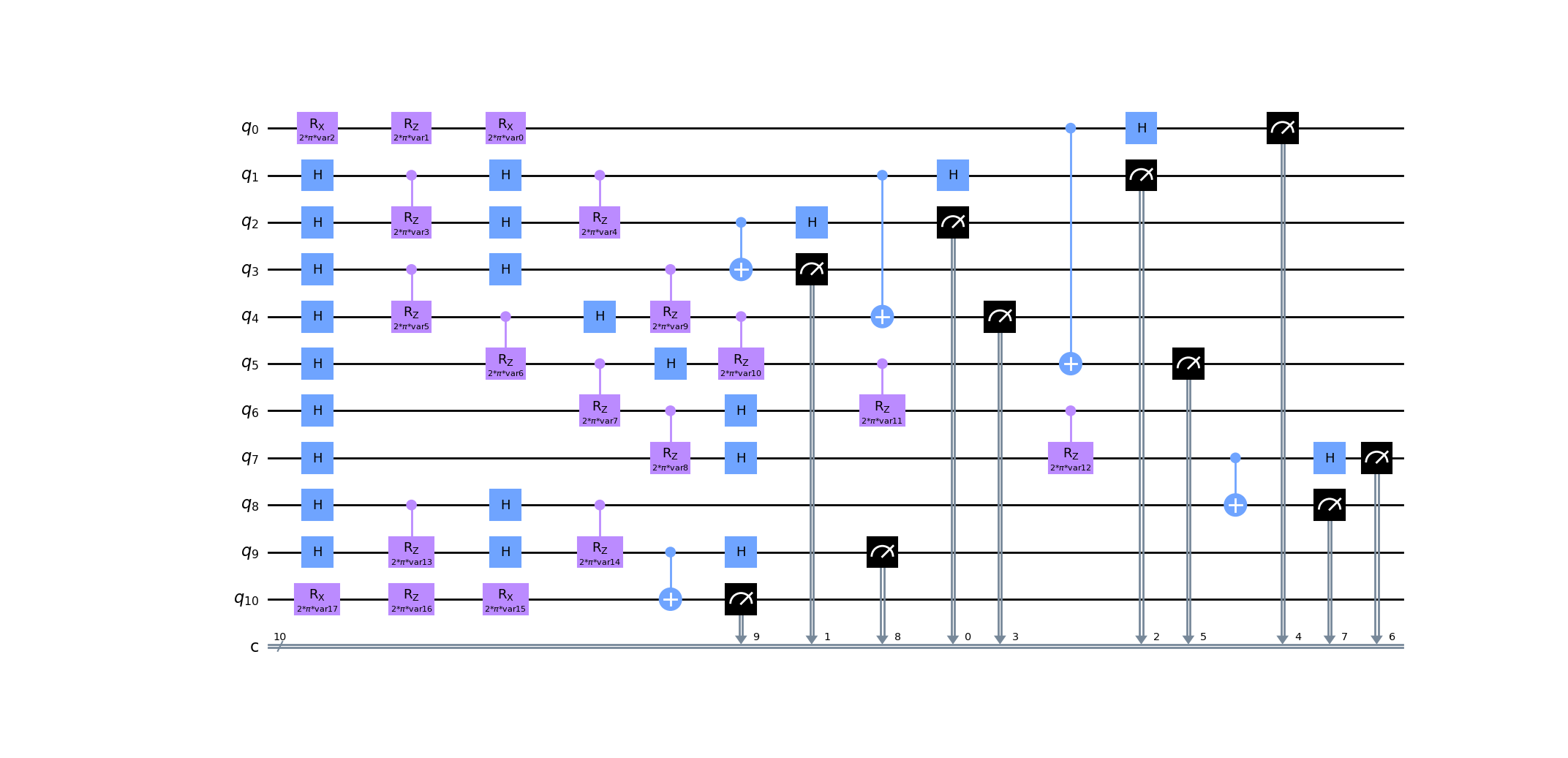}
\caption{Quantum circuit for the sentence ``John walks in the park'' in \texttt{qiskit} format. Noun and sentence types have been assigned 1 qubit each.}
\label{fig:qcircuit}
\end{figure}

Up to this point of the pipeline, a sentence is still represented as a string diagram, independent of any low-level decisions such as tensor dimensions or specific quantum gate choices. This abstract form can be turned into a concrete quantum circuit or tensor network by applying ans\"atze. An \textit{ansatz} can be seen as a map that determines choices such as the number of qubits that every wire of the string diagram is associated with and the concrete parameterised quantum states that correspond to each word. In \lambeq, ans\"atze can be added by extending one of the classes \texttt{TensorAnsatz} or \texttt{CircuitAnsatz}, depending on the type of the experiment. For the quantum case, the library comes equipped with the class \texttt{IQPAnsatz}, which turns the string diagram into a standard IQP circuit\footnote{Instantaneous Quantum Polynomial -- a circuit which interleaves layers of Hadamard gates with diagonal unitaries \citep{HavlivcekEtAl_2019_SupervisedLearning}.}. For instance, the code in Example \ref{ex:circuit} produces the circuit in Figure \ref{fig:qcircuit} by assigning 1 qubit to the noun type and 1 qubit to the sentence type. 

In the case of a classical experiment, parameterising with the \texttt{TensorAnsatz} class with $d_n=4$ for the base dimension of the noun space, and $d_s=2$ as the dimension of the sentence space, produces the following tensor network:

\ctikzfig{tensornet-john}

\begin{figure}[h!]
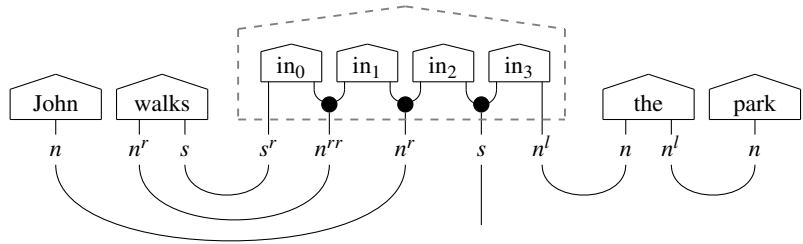
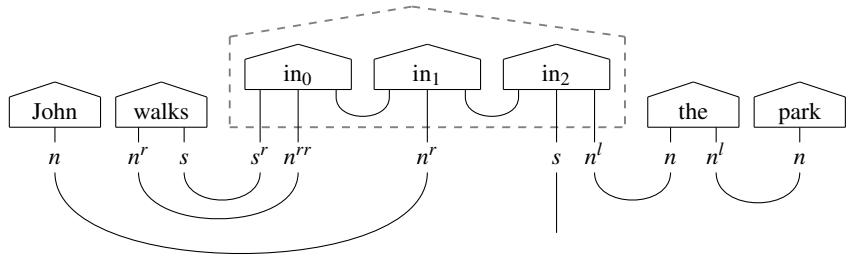

\centering
\begin{subfigure}{0.9\textwidth}
\ctikzfig{spider_ansatz}
\caption{\texttt{SpiderAnsatz} (maximum tensor order = 2).}
\label{fig:spider_ansatz}
\end{subfigure}
\begin{subfigure}{0.9\textwidth}
\ctikzfig{mps_ansatz}
\caption{\texttt{MPSAnsatz} (maximum tensor order = 3).}
\end{subfigure}
\caption{String diagrams using matrix product states.}
\label{fig:mpss}
\end{figure}

Note that in classical experiments of this kind, the tensors associated with certain words, such as conjunctions, can become extremely large. In some cases, the order of these tensors can be 12 or even higher ($d^{12}$ elements, where $d$ is the base dimension), which makes efficient execution of the experiment impossible. In order to address this problem, \lambeq~includes ans\"atze for converting tensors into various forms of \textit{matrix product states} (MPSs). Figure \ref{fig:mpss} shows the output of Example \ref{ex:tensor_ansatz}, where the original tensors have been split into smaller tensors linked with spiders (\texttt{SpidersAnsatz}) (a) or standard contractions (\texttt{MPSAnsatz}) (b).

\subsection{Optimisation/training}
\label{sec:training}

Although \lambeq~does not yet include a native machine learning module, it seamlessly collaborates with standard ML and optimisation libraries such as PyTorch and JAX. The documentation of the toolkit includes Jupyter notebooks that demonstrate a variety of classical and quantum pipelines (see also Section \ref{sec:practice}).

\section{lambeq in practice}
\label{sec:practice}

In this section we apply \lambeq~to a simple sentence classification task, demonstrating its usage in four different scenarios: A classical pipeline, where the sentences are encoded as tensor networks; a classical simulation of a quantum pipeline, where sentences are encoded as quantum circuits but the optimisation and training takes place on classical hardware; a shot-based simulation of a quantum pipeline, \textit{without} using a noise model, running on \texttt{qiskit}'s Aer simulator; and a noisy shot-based simulation of a quantum pipeline, again on Aer simulator. The code for the experiments can be found in the \texttt{examples} folder of the toolkit's documentation in the form of Jupyter notebooks. 

For our experiments we use the meaning classification dataset from \cite{lorenz2021qnlp}, in which the goal is to classify simple sentences (such as ``skillful programmer creates software'' and ``chef prepares delicious meal'') into two categories, food or IT. The dataset consists of 130 sentences created using a simple context-free grammar. We proceed to provide details about the parameterisation and results of the four experiments.

\begin{figure}
\centering
\includegraphics[scale=0.6]{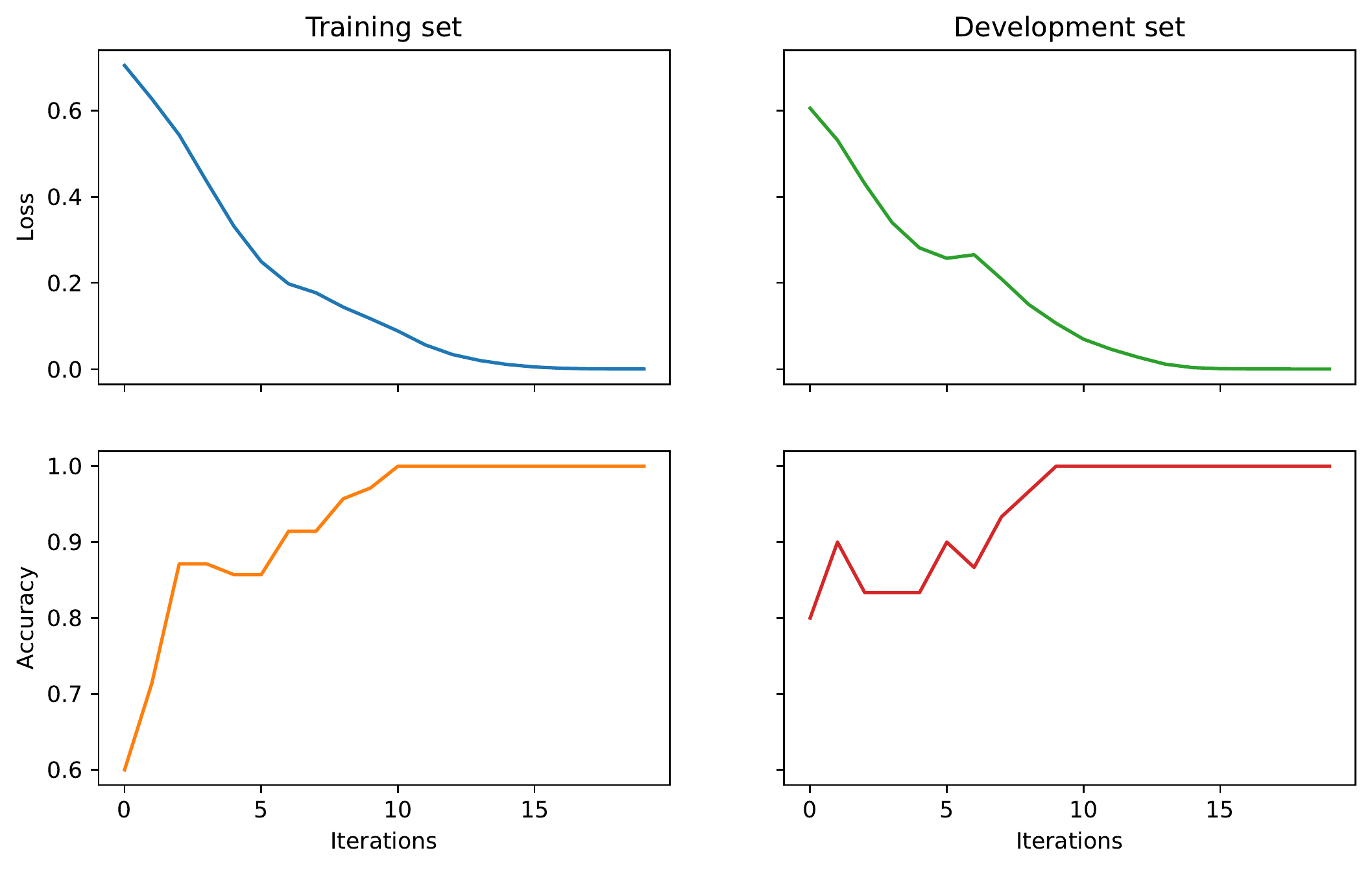}
\caption{Classical pipeline results.}
\label{fig:classical_plots}
\end{figure}

\paragraph{Classical pipeline} The first step in all the experiments is to convert the sentences into string diagrams with the \texttt{ccg2discocat} module. For the classical pipeline, we chose to apply \texttt{SpiderAnsatz} (Figure \ref{fig:spider_ansatz}), assigning a dimensionality of 2 to both the noun and sentence spaces. We use the PyTorch backend of Google's tensor network to perform the  training, with the Adam optimiser and standard binary cross entropy as the loss. Figure \ref{fig:classical_plots} gives plots for the accuracy and loss on the training and development sets. Perfect accuracy is also achieved on the test set. 

\begin{figure}
\centering
\includegraphics[scale=0.6]{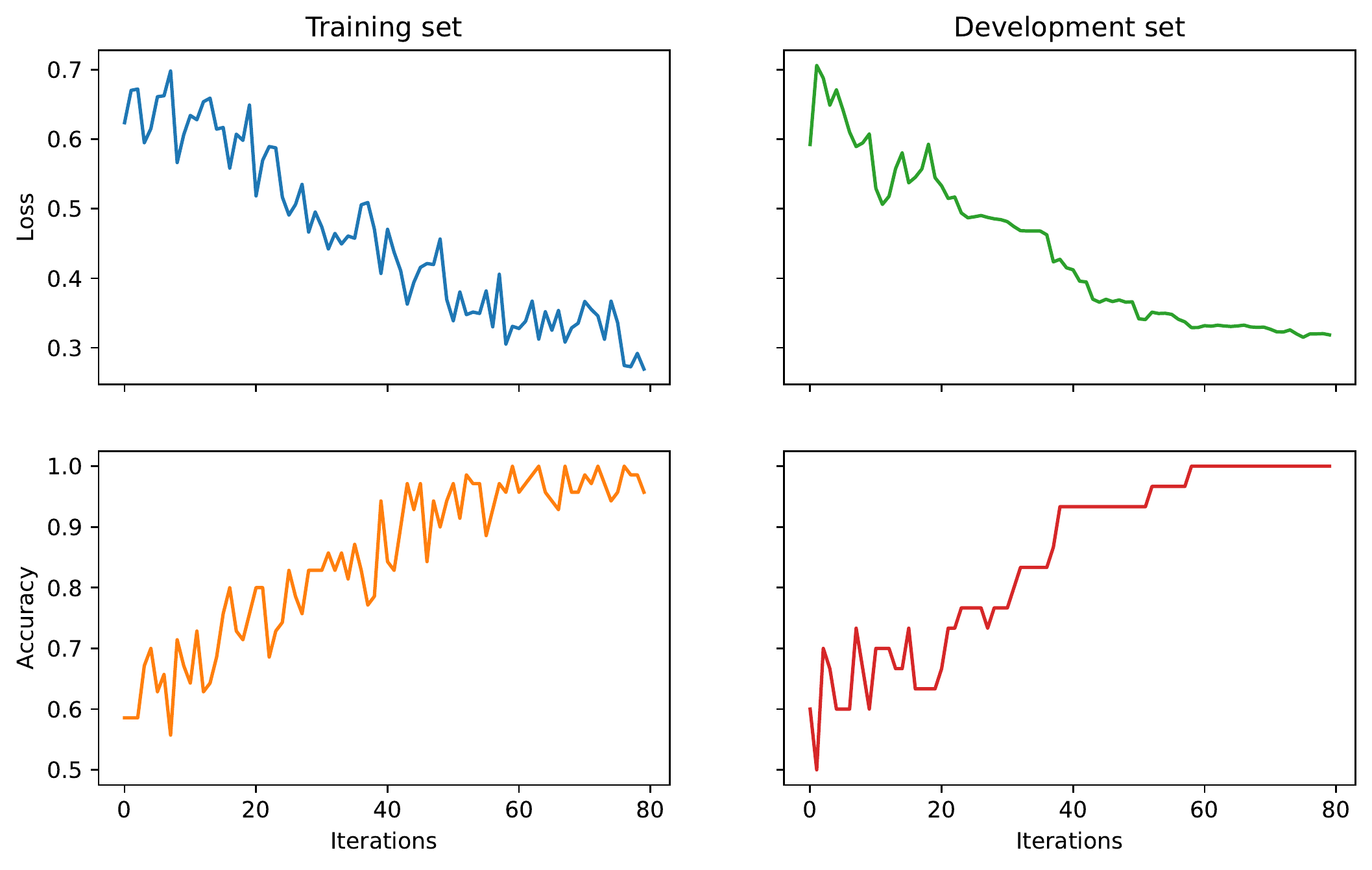}
\caption{Results for the ``classical'' simulation of a quantum pipeline.}
\label{fig:quan_sim_plots}
\end{figure}

\paragraph{Classical simulation of quantum pipeline} In this scenario the string diagrams are converted into quantum circuits using \texttt{IQPAnsatz}, with 1 qubit assigned to both the noun and sentence spaces. The number of IQP layers were set to 1. For optimisation we use a gradient-approximation technique, known as \textit{Simultaneous Perturbation Stochastic Approximation} (SPSA) \citep{spall1992multivariate}. The reason for this choice is that in a variational quantum circuit context like here, proper backpropagation requires some form of ``circuit differentiation'' that would in turn have to be evaluated on a quantum computer -- something  very costly from a practical perspective. SPSA provides a less effective but acceptable choice for the purposes of these experiments. The main prediction functions are ``just-in-time'' compiled with JAX in order to improve speed. Results are shown in Figure \ref{fig:quan_sim_plots}. Note that this configuration requires about 4 times more iterations than the previous one in order to converge; however this is eventually achieved. Once again, we obtain perfect accuracy on the test set. 

\begin{figure}
\centering
\includegraphics[scale=0.6]{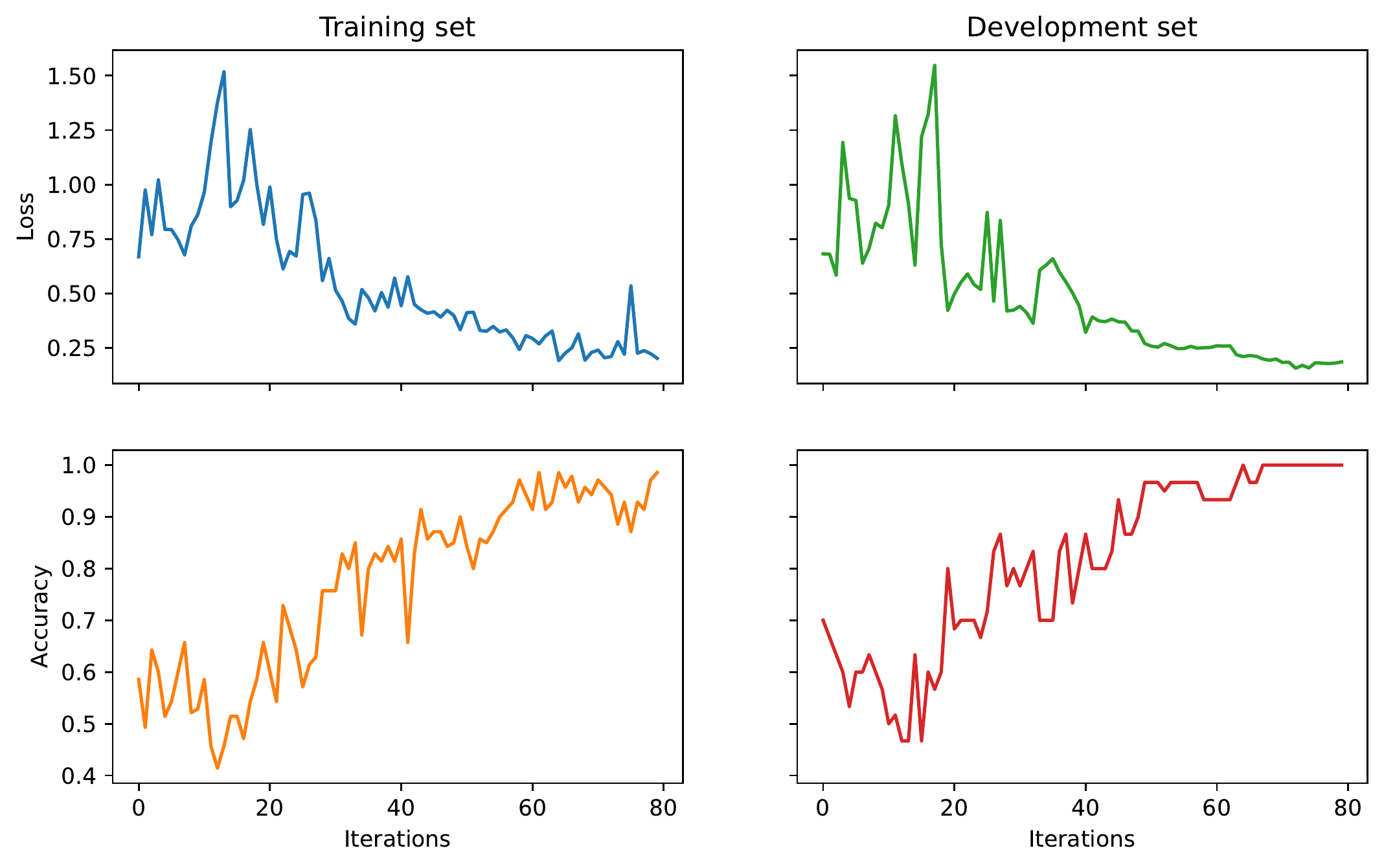}
\caption{Results for the noiseless shot-based simulation of a quantum pipeline.}
\label{fig:quan_em_plots}
\end{figure}

\paragraph{Noiseless shot-based simulation of quantum pipeline} For this experiment we repeat the configuration of the classical simulation pipeline, but this time we evaluate the model on \texttt{qiskit}'s Aer simulator that is available through \texttt{tket}'s Python interface, \texttt{pytket}. The simulation is performed without using a noise model. SPSA is again used as the optimisation algorithm. Figure \ref{fig:quan_em_plots} shows that while there is a lot of fluctuation and instability at the early stages of training, the model eventually converges successfully. Accuracy on the test set is again perfect, although it is worth noting that performance can vary considerably between different runs.

\begin{figure}
\centering
\includegraphics[scale=0.6]{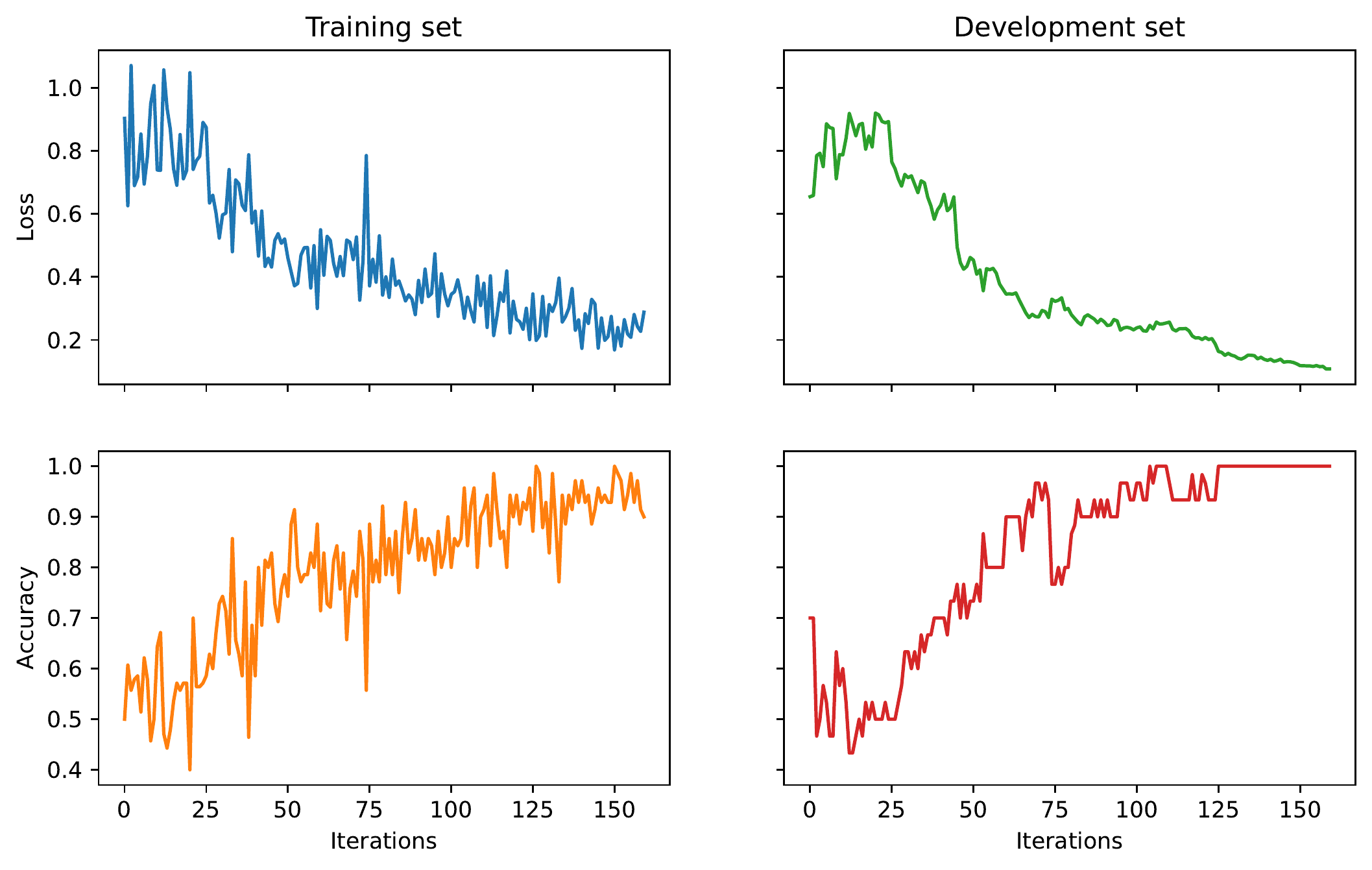}
\caption{Results for the noisy shot-based simulation of a quantum pipeline.}
\label{fig:quan_sim_noisy}
\end{figure}

\vspace{-0.3cm}
\paragraph{Noisy shot-based simulation of quantum pipeline} Finally, we use again the Aer simulator to run a noisy simulation, which is the closest we can get to an actual run on real quantum hardware. Excluding the noise model, the setting is identical with the one of the noiseless run. In Figure \ref{fig:quan_sim_noisy} we can see that even more iterations were required for the model to converge, due to the noisy nature of the simulation. However, perfect accuracy was achieved once more on the test set.

\subsection{Other uses of lambeq}
\label{sec:other_uses}

\begin{figure}
\centering
\includegraphics[scale=0.42, trim=0 32cm 0 0, clip]{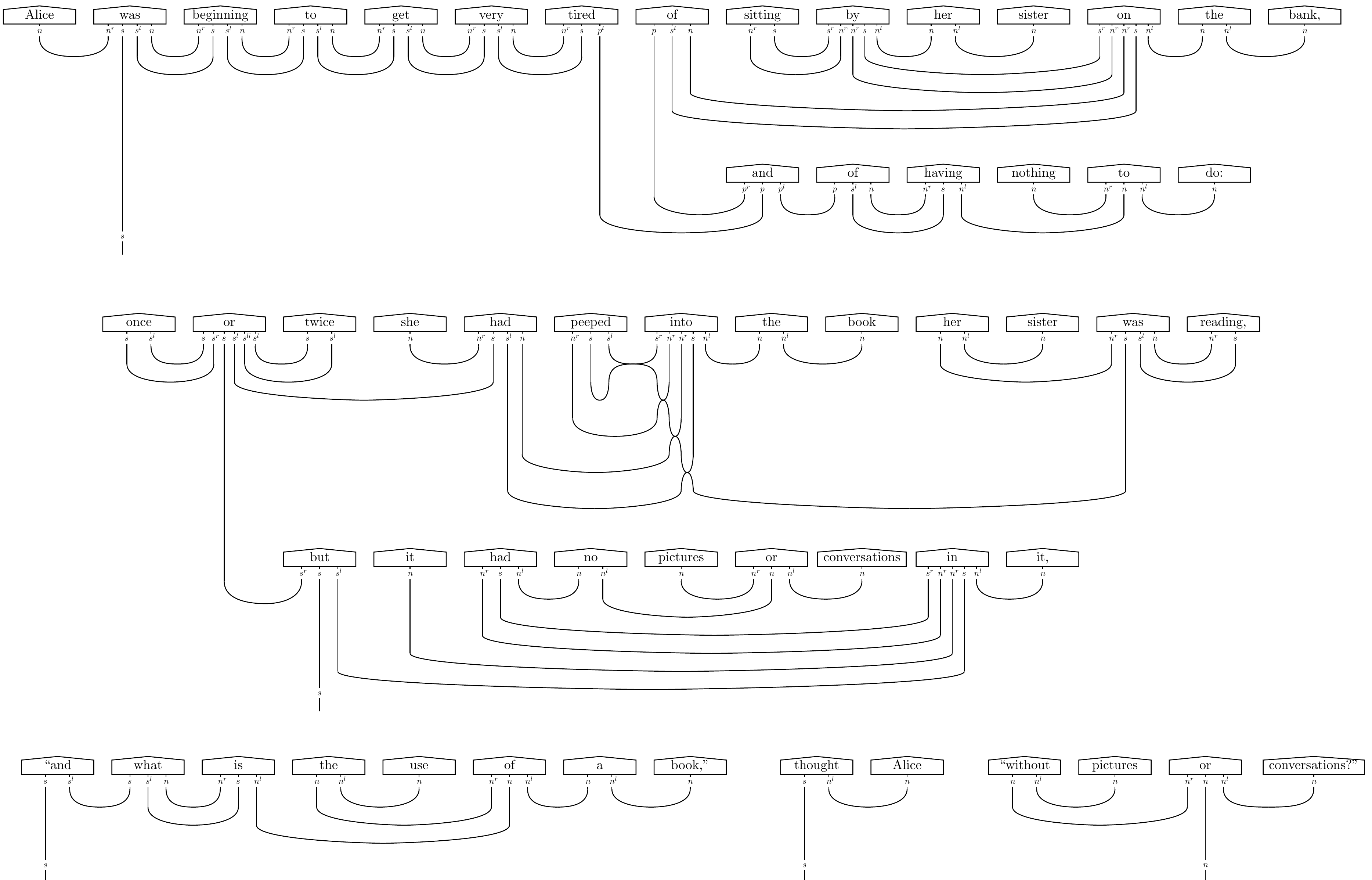}
\caption{The first few sentences of ``Alice in Wonderland'' in diagrammatic form. The resource was created using \lambeq's \texttt{ccg2discocat} module.}
\label{fig:alice}
\end{figure}

Below are some links to work that has previously been carried out by CQC's QNLP team using the toolkit, before this first official release:

\begin{itemize}
\item An early version of the toolkit has been used for the small-scale experiments on quantum hardware by \cite{lorenz2021qnlp}, implementing the pipeline of Figure \ref{fig:gen_pipeline}. 

\item A web tool\footnote{\url{https://qnlp.cambridgequantum.com/generate.html}} based on \lambeq's \texttt{ccg2discocat} module was made available in the summer of 2021, allowing the conversion of any sentence into a string diagram or quantum circuit, and supporting various output formats. 

\item \cite{yeung-kartsaklis} used \lambeq~in order to create the first large-scale DisCoCat resource including more than 3,000 string diagrams, parsing the book ``Alice in Wonderland'' into DisCoCat form\footnote{\url{https://qnlp.cambridgequantum.com/downloads.html}} (Figure \ref{fig:alice}).
\end{itemize} 

\section{Conclusions and future work}
\label{sec:future_work}

We introduced \lambeq, the first high-level open-source Python toolkit for quantum natural language processing. This first release includes abstractions and tools for implementing all the necessary stages of a pipeline that converts sentences into quantum circuits and tensor networks. We briefly outline our plans for the future. 

As mentioned before, providing more choices of compositional models and ans\"atze, for both quantum and classical experimental pipelines, is one of our top priorities and currently work in progress. Support for additional  parsers will also be included in one of the future versions of the package. Specifically, \lambeq~will be extended with the transformer-based CCG parser described in \cite{clark2021}, which currently significantly improves the state-of-the-art. 

A more ambitious direction we plan to follow is the introduction of a ``quantum-friendly'' optimisation/training module, providing easy access to machine learning algorithms, objectives, and techniques suitable to quantum hardware. This is a long-term goal that is more likely to be fulfilled in stages. A priority is to integrate into \lambeq~the diagrammatic differentiation features that were recently introduced in DisCoPy \citep{toumi2021}, taking one step closer to a fully automated quantum machine learning pipeline.

Another longer-term goal is the addition of functionality that supports discourse-related tasks at the paragraph or document level, according to the DisCoCirc model \citep{discocirc}. The idea there is to encode an entire document into a quantum circuit, where nouns are dynamically modified by the interaction with functional words, such as verbs and adjectives, as the text flows from one sentence to another. 

Finally, one of the main directions of our team is the active support, maintenance and further contributions to the development of DisCoPy, \lambeq's low-level backend. 

\subsection*{Acknowledgements}

We would like to thank our colleagues in Cambridge Quantum Computing for their help and support. We are especially grateful to the Oxford QNLP team for all those valuable discussions, insights and exchange of ideas during the past year. The name of the toolkit is a tribute to the great Joachim Lambek, whom Bob Coecke had the pleasure to know personally, and even live in his house for some months as a postdoc in Montreal. In 2004 Bob was presenting categorical quantum mechanics at the McGill category theory seminar; when Lambek saw quantum teleportation with diagrams, he pointed to the screen and said: ``This is grammar!''.  And right he was!  

\vspace{0.2cm}
We acknowledge the use of IBM Quantum services for this work. The views expressed are those of the authors, and do not reflect the official policy or position of IBM or the IBM Quantum team. 

\bibliographystyle{plainnat}
\bibliography{ref}
\newpage

\appendix
\section{Appendix}

\begin{example}{Using \texttt{ccg2discocat}.}
\begin{lstlisting}
from lambeq.ccg2discocat import DepCCGParser

# Parse the sentence and convert it into a string diagram
depccg_parser = DepCCGParser()
diagram = depccg_parser.sentence2diagram('John gave Mary a flower')

diagram.draw()
\end{lstlisting}
\label{ex:ccg2discocat}
\end{example}

\begin{example}{\texttt{SpidersReader} and \texttt{CupsReader}.}
\begin{lstlisting}
from lambeq.reader import cups_reader, spiders_reader

sentence = 'John gave Mary a flower'

# Create string diagrams based on spiders and cups reader
spiders_diagram = spiders_reader.sentence2diagram(sentence)
cups_diagram = cups_reader.sentence2diagram(sentence)

spiders_diagram.draw()
cups_diagram.draw()
\end{lstlisting}
\label{ex:readers}
\end{example}


\begin{example}{\texttt{CCGBankParser}}
\begin{lstlisting}
from lambeq.ccg2discocat import CCGBankParser

# Replace path below with the root folder of CCGBank in your system.
# The sections must be located in <root>/data/AUTO
parser = CCGBankParser(root='/ccgbank')
diagrams_section_00 = parser.section2diagrams(section_id=0)

\end{lstlisting}
\label{ex:ccgbank}
\end{example}

\begin{example}{Rewriting.}
\begin{lstlisting}
from lambeq.ccg2discocat import DepCCGParser
from lambeq.rewrite import Rewriter

# Parse the sentence
diagram = depccg_parser.sentence2diagram('John walks in the park')

# Apply rewrite rule for prepositional phrases
rewriter = Rewriter(['prepositional_phrase'])
rewritten_diagram = rewriter(diagram)

rewritten_diagram.draw()
\end{lstlisting}
\label{ex:rewrite}
\end{example}

\begin{example}{Generating a quantum circuit.}
\begin{lstlisting}
from lambeq.ccg2discocat import DepCCGParser
from lambeq.circuit import IQPAnsatz
from lambeq.core.types import AtomicType

# Define atomic types
N = AtomicType.NOUN
S = AtomicType.SENTENCE

# Get a string diagram
depccg_parser = DepCCGParser()
diagram = depccg_parser.sentence2diagram('John walks in the park')

# Convert string diagram to quantum circuit
ansatz = IQPAnsatz({N: 1, S: 1}, n_layers=1)
discopy_circuit = ansatz(diagram)
discopy_circuit.draw()

# Convert to tket form
tket_circuit = discopy_circuit.to_tk()
\end{lstlisting}
\label{ex:circuit}
\end{example}

\begin{example}{Tensor ans\"atze.}
\begin{lstlisting}
from lambeq import ccg2discocat
from lambeq.tensor import TensorAnsatz, MPSAnsatz, SpiderAnsatz
from lambeq.core.types import AtomicType
from discopy import Dim

# Define atomic types
N = AtomicType.NOUN
S = AtomicType.SENTENCE

# Parse the sentence
depccg_parser = ccg2discocat.DepCCGParser()
diagram = depccg_parser.sentence2diagram('John walks in the park')

# Standard tensor diagram (no splitting of tensors)
std_diagram = TensorAnsatz({N: Dim(4), S: Dim(2)})(diagram)

# MPS tensor diagram
mps_diagram = MPSAnsatz({N: Dim(4), S: Dim(2)}, bond_dim=4)(diagram)

# Spiders diagram
spider_diagram = SpiderAnsatz({N: Dim(4), S: Dim(2)})(diagram)

std_diagram.draw()
mps_diagram.draw()
spider_diagram.draw()
\end{lstlisting}
\label{ex:tensor_ansatz}
\end{example}

\end{document}